\documentclass[10pt,twocolumn,letterpaper]{article}

\usepackage{iccv}
\usepackage{times}
\usepackage{epsfig}
\usepackage{graphicx}
\usepackage{amsmath}
\usepackage{amssymb}
\usepackage{graphicx}
\usepackage{times}
\usepackage{epsfig}
\usepackage{subcaption}
\usepackage{multirow}
\usepackage{multicol}
\usepackage{floatrow}
\usepackage{algorithm}
\usepackage{algorithmic}
\usepackage{booktabs}
\usepackage{mathrsfs}

\usepackage{array}
\newcolumntype{I}{!{\vrule width 3pt}}
\newlength\savedwidth
\newcommand\whline{\noalign{\global\savedwidth\arrayrulewidth
                           \global\arrayrulewidth 2pt}%
                  \hline
                  \noalign{\global\arrayrulewidth\savedwidth}}
\newlength\savewidth
\newcommand\shline{\noalign{\global\savewidth\arrayrulewidth
                           \global\arrayrulewidth 0.5pt}%
                  \hline
                  \noalign{\global\arrayrulewidth\savewidth}}
                  
\newcommand{\green}[1]{\textcolor[RGB]{96,177,87}{#1}}

\def\eg{\emph{e.g., }}

\def\ie{\emph{i.e., }}

\usepackage[pagebackref=true,breaklinks=true,letterpaper=true,colorlinks,bookmarks=false]{hyperref}

\iccvfinalcopy 

\def\iccvPaperID{6585} 

\ificcvfinal\fi

\begin{document}

\title{Polygon-free: Unconstrained Scene Text Detection with Box Annotations}

\author{Weijia Wu$^1$,
Enze Xie$^2$,
Ruimao Zhang$^3$,
Wenhai Wang$^4$,
Ping Luo$^2$,
Hong Zhou$^1$\\
$^1$Zhejiang University\\
$^2$University of Hong Kong Hong Kong SAR\\
$^3$The Chinese University of Hong Kong, Shenzhen China 
\\
$^4$Shanghai AI Laboratory
}

\maketitle
\ificcvfinal\thispagestyle{empty}\fi

\begin{abstract}
 
Although a polygon is a more accurate representation than an upright bounding box for text detection, the annotations of polygons are extremely expensive and challenging. Unlike existing works that employ fully-supervised training with polygon annotations, this study proposes an unconstrained text detection system termed Polygon-free~(PF),
in which most existing polygon-based text detectors~(\eg{PSENet~\cite{psenet},DB~\cite{liao2020real}}) are trained with only upright bounding box annotations. 
Our core idea is to transfer knowledge from synthetic data to real data to enhance the supervision information of upright bounding boxes.
This is made possible with a simple segmentation network, namely Skeleton Attention Segmentation Network~(SASN), that includes three vital components~(\ie{channel attention, spatial attention and skeleton attention map}) and one soft cross-entropy loss.

Experiments demonstrate that the proposed Polygon-free system can combine general detectors~(\eg{EAST, PSENet, DB}) to yield surprisingly high-quality pixel-level results with only upright bounding box annotations on a variety of datasets~(\eg{ICDAR2019-Art, TotalText, ICDAR2015}). For example, without using polygon annotations, PSENet achieves an 80.5\% F-score on TotalText~\cite{totaltext} ~(vs. 80.9\% of fully supervised counterpart), \textbf{31.1\%} better than training directly with upright bounding box annotations, and saves \textbf{80\%+} labeling costs. 
We hope that PF can provide a new perspective for text detection to reduce the labeling costs.
The code can be found at \href{https://github.com/weijiawu/Unconstrained-Text-Detection-with-Box-Supervisionand-Dynamic-Self-Training}{\color{blue}{$\tt github.com/weijiawu/Polygon-free$}}.
\end{abstract}

\begin{figure}[t]
\begin{center}
\includegraphics[width=1\textwidth]{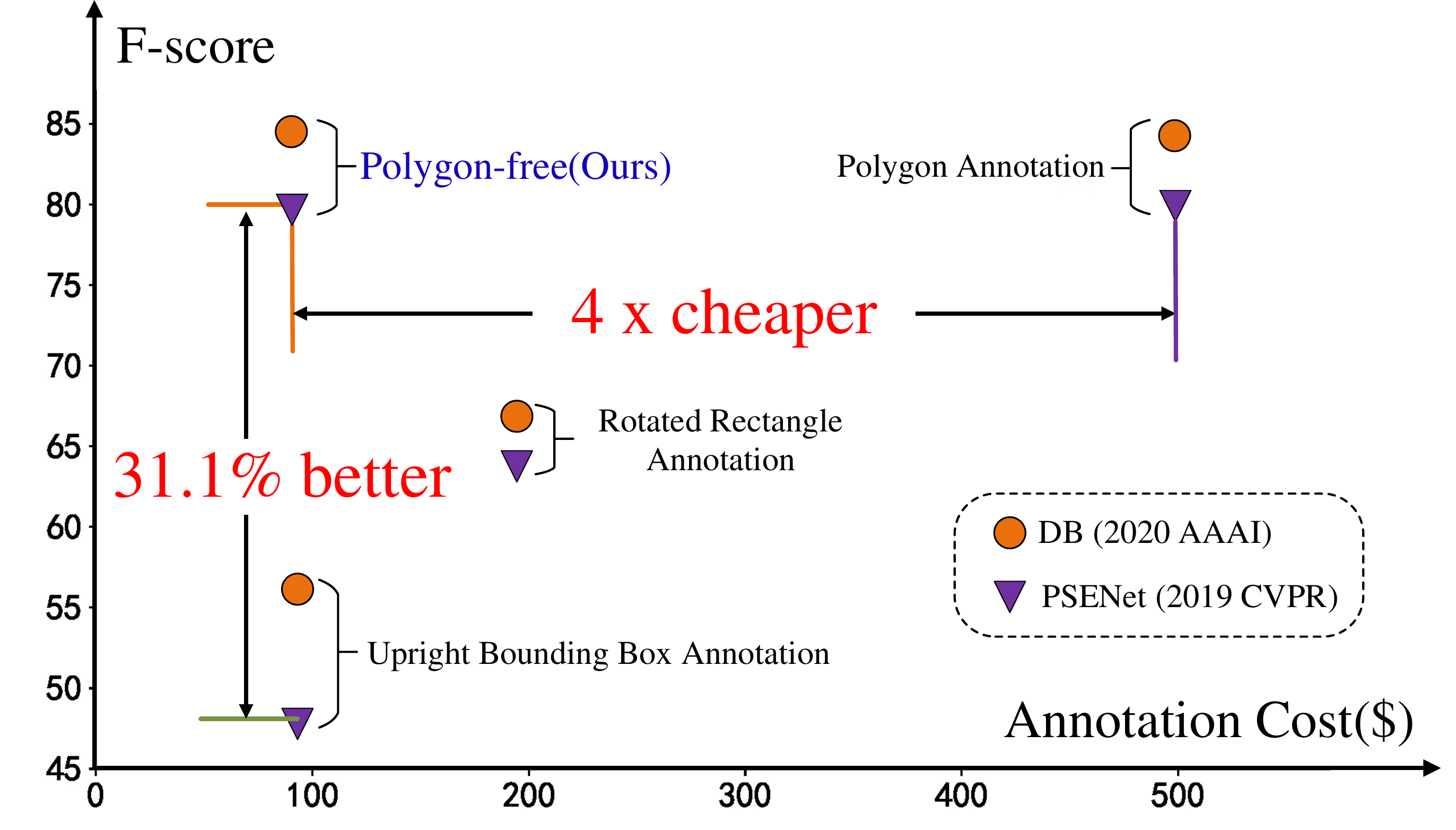}
\caption{\textbf{The performance and annotation cost for PSENet~\cite{psenet} and DB~\cite{liao2020real} on TotalText~\cite{totaltext}}. PSENet with Polygon-free is $31.1\%$ better than training directly with upright bounding box annotation, $4\times $ times cheaper than that with polygon annotation. Note that the rough cost evaluation of Total-Text is obtained by Amazon Mechanical Turk.}
\vspace{-0.5cm}
\label{cost}
\end{center}
\end{figure}

\section{Introduction}
\begin{figure}[t]
\begin{center}
\includegraphics[width=1\textwidth]{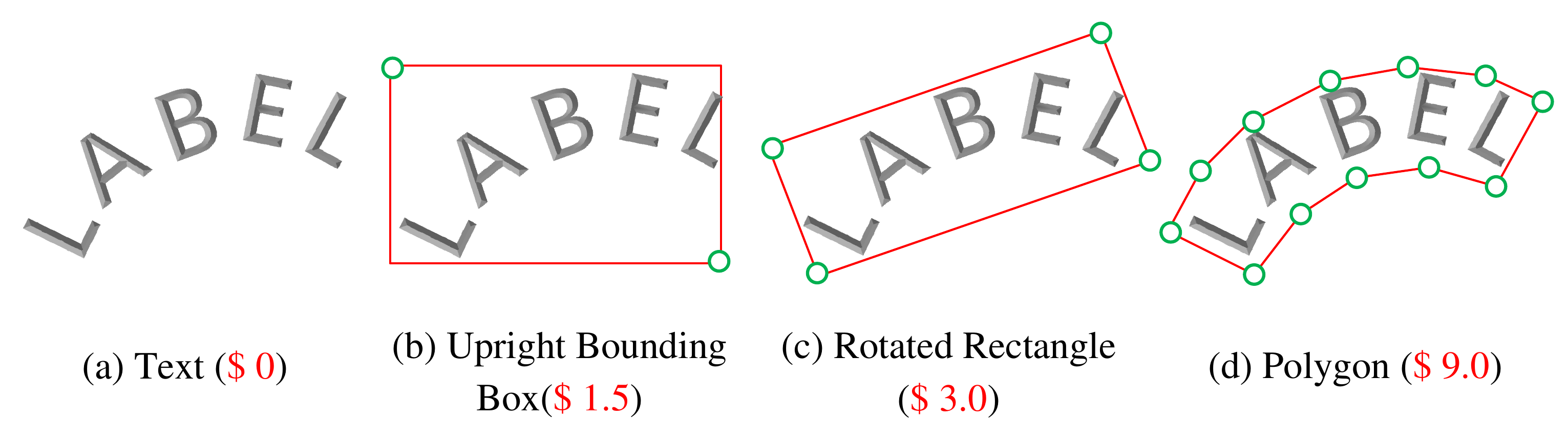}
\caption{\textbf{Comparisons of annotation cost}. Annotation of more points is more expensive~(the price of 100 text instances). Note that the data cost information is obtained from Amazon Mechanical Turk~(MTurk).}
\vspace{-0.5cm}
\label{annotation}
\end{center}
\end{figure}

Scene text detection~\cite{tian2016detecting,liao2017textboxes,wu2019textcohesion,zhou2017east,textsnake,masktextspotter,wu2020synthetic,wu2022end,li2021contrastive}, which aims to locate texts in the wild, has achieved much attention in recent years because of its numerous applications, such as instant translation, image retrieval, scene parsing. Unlike other general objectives, scene text usually cannot be described accurately by the upright bounding box~(see the example in Fig.~\ref{annotation}~(b)), because of the diverse shapes, so most detectors using an upright bounding box annotation only achieve F-scores of below $60\%$, such as the $49.6\%$ F-score for PSENet\cite{psenet} on TotalText~\cite{totaltext}. Recently, to achieve improved performance, most scene text detection methods~\cite{psenet,textsnake,wang2019efficient} have utilized polygon annotation with many coordinates~(see the example in Fig.~\ref{annotation}~(d)) to capture texts with different shapes. Although the polygon annotations are more accurate than the upright bounding box annotations, the labeling cost of polygons is extremely high, limiting the wide use of this method in large-scale real-world applications~\cite{wu2021bilingual}. According to the annotation report of the large-scale ICDAR2019-LSVT~\cite{sun2019chinese} dataset, 50K polygon annotations require the work of 55 personnel over five weeks, \textit{i.e.,} 11,000 man-hours, which is time-consuming and frustrating. 
By contrast, the upright bounding box annotations are more economical, and are approximately 4$\times$ cheaper than polygons annotations~\cite{totaltext,yuliang2017detecting}, \eg{saving 80\% + annotation cost on TotalText~\cite{totaltext}, as shown in Fig.~\ref{cost}}. This cost gap becomes larger for the large-scale benchmarks such as ICDAR2019-LSVT~\cite{sun2019chinese} and ICDAR2019-Art~\cite{chng2019icdar2019}. 
Therefore, we aim to eliminate this obstacle by training pixel-level text detectors using upright box annotations only.

A few previous works such as BoxSup~\cite{dai2015boxsup} and Box2Seg~\cite{kulharia2020box2seg} attempted to train semantic or instance segmentation network with pseudo label from MCG~\cite{pont2016multiscale} or GrabCut~\cite{rother2004grabcut} based on upright bounding box annotations. One of the drawbacks of these approaches is the complicated training pipeline and a high number of hyper-parameters. Moreover, none of these methods are able to show strong weakly supervised performance. Most importantly, these methods are designed for the detection of general objects rather than for text detection. To address the text data cost issue, we first propose a simple, unconstrained system termed Polygon-free~(PF) for training text detector with upright bounding box annotated data.
The core idea is to transfer knowledge from low-cost synthetic data to real data to enhance the supervision information of upright bounding boxes by Attention Mechanism as inspired by the following observations.

Compared with polygon annotations, the upright bounding box is much less expensive but contains less pixel information for efficient supervision.
Thus, effectively utilizing the upright bounding box annotations and boosting the text detection performance becomes critical in this case.
An alternative approach is to utilize synthetic text data~\cite{synthtext,long2020unrealtext} that are largely available from the virtual world, and the ground truth can be freely and automatically generated.
However, many previous works~\cite{synthtext} have shown that training directly with synthetic data degrades the performance on real data due to a phenomenon known as ``domain shift"~(\eg{58.0\% for EAST directly training on SynthText and testing on ICDAR2015}). 
Unlike the existing works, motivated by the attention mechanism studies~\cite{woo2018cbam,chen2017sca} and BoxSup~\cite{dai2015boxsup}, we propose a Skeleton Attention Segmentation Network~(SASN), and carefully design a Skeleton Attention Module based on channel attention and spatial attention to reduce the domain shift, help the network learn domain-invariant features with stronger representation power for text in a prior upright bounding box. 
Besides, considering the particular geometry of texts, we argue that it is more important to focus on the skeleton than on other regions. Therefore, we introduce a soft attention weight map called skeleton map and corresponding soft cross-entropy loss to enhance the representation power of the text skeleton.

To make it suitable for all detectors, the whole Polygon-free is divided into two steps: we \textbf{first} apply the synthetic data with character-level annotations to pre-train the SASN.
By exploiting the text data with various geometric information~(\eg{ the straight and the curved data }) in the training phase, SASN can effectively capture the changes of the text appearance in different scenes with the channel and spatial attention.
\textbf{Then} upright bounding box annotations are used to crop the text region in the real images, which are fed into SASN to generate high-quality polygon-like pseudo labels.
By splicing all of these local pseudo labels, the global one would be obtained 
and applied to train any text detectors.
As shown in Fig.~\ref{cost}, by using DF, PSENet~\cite{psenet} achieves $4\times$ lower data cost compared with that of using polygon annotation, and obtains $31.1\%$ F-score improvements compare with that of directly using upright bounding box annotation on Total-Text~\cite{totaltext}. The main contributions are three folds:

(1) We first demonstrate a simple, unconstrained Polygon-free system that can train most existing text detectors with only upright bounding box annotations. This means that general detectors~(\eg{PSENet~\cite{psenet}, DB~\cite{liao2020real}}) can be trained by upright bounding box annotations with no modification to the network itself.

(2) We introduce a Skeleton Attention Segmentation Network composed of three vital components (\ie{Spatial Attention, Channel Attention and Skeleton Attention}) and one soft cross-entropy loss, which capturing stronger text representation and bridging the domain gap between synthetic data and real data.

%
(3) Polygon-free attains excellent pixel level text detection performances on several \textbf{large-scale} datasets~(\eg{ICDAR2019-LSVT~\cite{sun2019chinese} with \textbf{50k} images and ICDAR2019-Art~\cite{chng2019icdar2019} with \textbf{10k+} curved text images}). With no polygon annotations used in training, PSENet achieves \textbf{77.6\%} F-score on ICDAR2019-LSVT, \textit{even outperforming its fully supervised counterpart~(\ie{77.4\% F-score})}. The cost reduction for \textbf{50k} polygon annotation requiring \textbf{11,000} man-hours is highly significant.

\begin{figure*}[t]
\begin{center}
\includegraphics[width=1\textwidth]{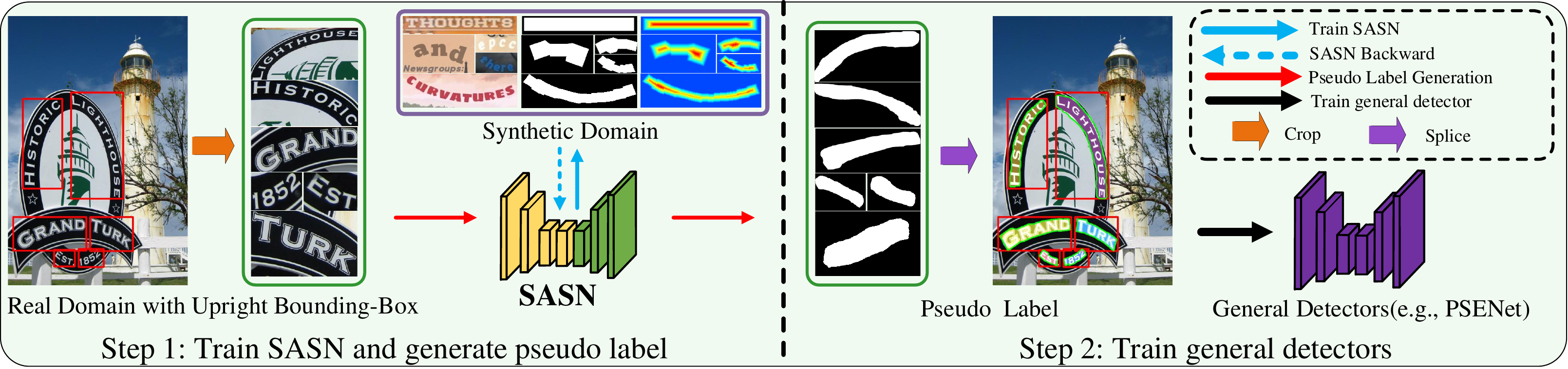}
\caption{\textbf{Illustration of the whole Polygon-free pipeline}. Training of detectors with only upright bounding box includes two steps: (1) Train SASN~(\ie{Blue 
arrows}) with synthetic data and inference on real data~(\ie{Red 
arrows}) to generate polygon-like pseudo label, details shown in Fig.~\ref{fig:box}. (2) Train general detectors~(\eg{PSENet~\cite{psenet}}) with the pseudo label.}

\label{fig:cost}
\end{center}
\vspace{-0.5cm}
\end{figure*}

\section{Related Work}
\subsection{Supervised Text Detection}

Scene text detection has achieved remarkable progress in the deep learning era. 
Previous methods~\cite{liao2017textboxes,zhang2016multi} focused on horizontal or multi-oriented text detection. CTPN~\cite{tian2016detecting} adopted Faster RCNN~\cite{ren2015faster} and modified RPN to detect texts.
EAST~\cite{zhou2017east} used FCN~\cite{FCN} to predict the text score map, distance map and angle map.
Recent works focused on curved text detection~\cite{baek2019character,xu2019textfield,masktextspotter}. 
TextSnake~\cite{textsnake} modeled curved text instance as a series of ordered disks and a text center line.
PSENet~\cite{psenet} and PAN~\cite{wang2019efficient} treat text instances as kernels with different scales and reconstruct the whole text instance in the post-processing.
There are a few previous works~\cite{Wetext} concerning weakly supervised text detection.
WordSup\cite{WordSup} trains a character detector by exploiting word annotations in rich large-scale real scene text datasets.

\subsection{Box-supervised Segmentation}
\textbf{Semantic Segmentation.} To alleviate the expensive data cost problem, a few works attempted to obtain semantic mask using upright box annotations. For example, BoxSup~\cite{dai2015boxsup} train an FCN with the pseudo labels from MCG~\cite{pont2016multiscale}, and an iterative training algorithm is used to refine the semantic masks. Box2Seg~\cite{kulharia2020box2seg} employs the pseudo label generated by GrabCut~\cite{rother2004grabcut} to supervise the training of the mask prediction model. At the same time, a per-class attention map is also predicted to make the per-pixel cross-entropy loss focus on the foreground pixels.

\textbf{Instance Segmentation.} Compared to semantic segmentation, utilizing box annotations to supervise instance segmentation networks is more rare. Similar to the methods for semantic segmentation, SDI~\cite{khoreva2017simple} also utilizes the pseudo label generated by MCG~\cite{pont2016multiscale}; however, it employs GrabCut-like algorithms to generate training labels from given bounding boxes, instead of modifying the segmentation convnet training procedure or using recursive training. Recently, BoxInst~\cite{tian2020boxinst} train the instance segmentation framework with box supervision by introducing two loss terms for CondInst~\cite{tian2020conditional}. However, the above works all have one drawback, the above works are designed for common object detection and segmentation, which cannot directly be applied to text detection. Our Polygon-free can fill the gap for box-supervised text detection.

\subsection{Attention Mechanism}
One important property of a human visual system is that we know ``what'' and ``where'' to focus our attention in an image. Similarly, attention enables the artificial model to focus only on the important data. The classic works based on attention, BAM~\cite{woo2018cbam} and CBAM~\cite{woo2018cbam} increase the
accuracy of the classifier by utilizing both 1D channel and 2D spatial self-attention maps. CBAM learns better feature representation by refining features with attention maps obtained from two different dimensions: channel and spatial. BAM constructs hierarchical attention at bottlenecks, and it is trainable in an end-to-end manner jointly with any feed-forward models. Considering the particular geometry of texts, we propose a novel skeleton attention map to replace the attention map in CBAM and carefully design a Skeleton Attention Segmentation Network with the channel and spatial attention to bridge the domain shift.

\section{Approach}

\subsection{Polygon-free System}\label{box}
\begin{figure*}[!t]
\begin{center}

\includegraphics[width=1\textwidth]{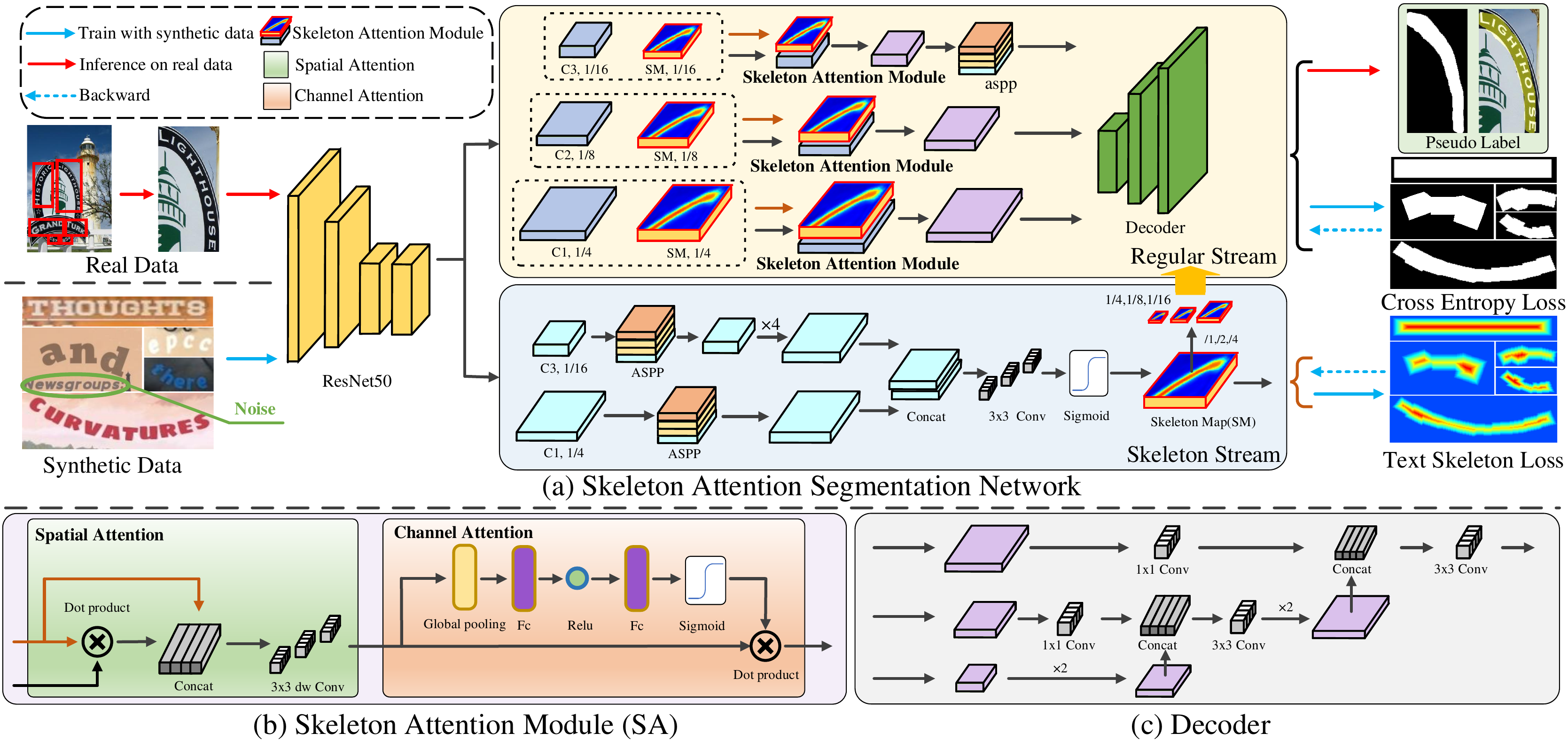}
\caption{\textbf{The network architecture of Skeleton Attention Segmentation Network.} (a) SASN is composed of two streams: regular stream and skeleton stream. (b) Skeleton Attention Module is composed of channel attention and spatial attention, which refine the input feature map by weighting with text skeleton map. (c) The detailed structure of the Decoder.}
\label{fig:box}
\end{center}
\end{figure*}

Inspired by the classic weakly-supervised work BoxSup~\cite{dai2015boxsup} that segmenting with upright bounding boxes, we propose an unconstrained weakly supervised method, termed Polygon-free~(PF), which can achieve competitive accuracy with low-cost upright bounding box annotations for text detection. As shown in Fig.~\ref{fig:cost}, PF utilizes a novel proposed Skeleton Attention Segmentation Network (SASN), which is shown in Fig.~\ref{fig:box}, to generate polygon pseudo labels based on the given upright bounding box annotations.
This process consists of two steps:
(1) In the first step (see the blue arrows in Fig.~\ref{fig:cost}), we train SASN with almost free synthetic data based on character annotation. 
And then, the box-level annotations are utilized to crop the real image, which are fed into the SASN for generating polygon-liked pseudo labels~(see red arrows in Fig.~\ref{fig:cost}). 
(2) By splicing all of the local pseudo labels, the global pseudo label is obtained. In this way, upright bounding box annotations can be converted to high-quality polygon pseudo labels. 
General detectors~( \eg{ PSENet~\cite{psenet}} ) trained on these pseudo labels can achieve almost the same performance as those trained on original polygon annotations.

\subsection{Skeleton Attention Segmentation Network}
\textbf{A Probabilistic Perspective for Skeleton Attention.}
Although synthetic data such as SynthText~\cite{synthtext} can be automatically generated with diversified appearance, the model trained with only synthetic data generally cannot obtain satisfactory performance on real scenes due to the existence of domain shift, \eg{$58.0\%$ F-score for EAST~\cite{zhou2017east} directly training on SynthText~\cite{synthtext} and testing on ICDAR2015~\cite{karatzas2015icdar}}. Compared with training on original training set of ICDAR2015~(\ie{$76.4\%$ F-score}), there exists a performance gap $18.4\%$. 
The prior upright bounding box can provide much prior knowledge, but there are still severe domain shifts.
Here, we provide a probabilistic perspective for this problem. The segmentation problem can be viewed as learning the posterior $P(M|I)$, where $I$ refers to the image representation and $M$ is the predicted mask of text instances. Let us denote the joint distribution of benchmark for segmentation as $P(M, I)$, and use $P_S(M, I)$ and $P_R(M, I)$ to denote the synthetic domain joint distribution and the real domain joint distribution~(see the left in Fig.~\ref{fig:cost}), respectively. In the presence of a domain shift, $P_S(M , I) \neq P_R(M , I)$. Using the Bayes’s Formula, the joint distribution can be decomposed as:
\begin{equation}
\small
P(M , I) = P(M|I)P(I)\,,
\label{equation1}
\end{equation}
Similar to the classification problem~\cite{wang2018deep}, we make the covariate shift assumption that the conditional probability $P(M|I)$ is the same for the two domains, and the domain distribution shift is caused by the difference on the marginal distribution $P(I)$.
In the text segmentation task, the image representation $I$ is actually the feature map from the network. Therefore, to handle the domain shift problem, we try to enforce the marginal distribution from two domains to be the same~(\ie{$P_S(I) = P_T(I)$}) by learning better discriminative text features.
Inspired by the CBAM~\cite{woo2018cbam}, we carefully design the Skeleton Attention Module~(SAM) to learn domain invariant features, focusing on important features and suppressing unnecessary ones from two aspects: channel attention and spatial attention~(see the left notes and (b) in Fig.~\ref{fig:box}). 
In addition, strong noise interference also exists in real data. For example, two texts in one upright bounding box for which we only require the body text as the foreground, as illustrated in the Fig.~\ref{fig:box}~(a) with the green circle. 
Thus, considering the special geometry of texts, we argue that the skeleton is more important than the other regions, particularly in the case of existing other text interference. Therefore, we introduce a soft attention weight map termed skeleton map to enhance the representation power of the text skeleton.

\textbf{Network Architecture.} 
As presented in Fig.~\ref{fig:box}~(a), we use ResNet50~\cite{he2016deep} as the backbone network for the SASN, and extract three levels of features~( \ie $C1,C2,C3$ ) from different downsampled scales~( \ie $1/4,1/8,1/16$ ).
After that, the skeleton stream fuses C1 and C3 to predict the skeleton attention map, and the Atrous Spatial Pyramid Pooling module (ASPP) is used to enlarge the receive field. The skeleton attention map is downsampled to make it suitable for multi-scale feature maps.
At the same time, the regular stream refines multi-scale features~( \ie $C1,C2,C3$ ) by applying the Skeleton Attention Module.
Finally, $C1,C2,C3$ are fed into the Decoder, as shown in Fig.~\ref{fig:box}~(c). The C3 and C2 are first fused by up-sampling and concatenation. The fused feature map is further up-sampled to fuse with $C1$ in the same approach.

%

\subsection{Skeleton Attention Module}
\textbf{Spatial and Channel Attention.} Fig.~\ref{fig:box}~(b) illustrates the details of the two attention methods. Given an intermediate feature map $F\in \mathbb{R}^{C\times H \times W}$~(\ie{the orange input in Fig.~\ref{fig:box}~(b)}) and an skeleton map $F_{sm}\in \mathbb{R}^{C\times H \times W}$~(\ie{the black input in Fig.~\ref{fig:box}~(b)}), spatial attention is first utilized to focus on ``where" is the text skeleton by multiplication and concatenating. The spatial attention is computed as:
\begin{equation}
F^{'} = f^{3\times 3}([F \otimes F_{sm};F])\,,
\label{equation2}
\end{equation}
where $\otimes$ denotes element-wise multiplication, and $f^{3\times 3}$ represents a convolution operation with the filter size of $3\times 3$. Following the CBAM~\cite{woo2018cbam}, the channel attention is utilized to focus on ``what" is meaningful given an input image by exploiting the inter-channel relationship of the feature. Global pooling is first used to aggregate spatial information of the feature map~$F^{'}$. Then, the fully connected layer is used to connect different channels, as each channel of a feature map is considered as a feature detector~\cite{zeiler2014visualizing}. In short, a 1D channel attention map~$M_{c}\in \mathbb{R}^{C\times 1 \times 1}$ can be obtained by feeding the feature map~$F^{'}$ into the $GlobalPool$-$Fc$-$Relu$-$Fc$-$Sigmoid$ layers. The final refined output from SAM can be computed as: $F^{''} = M_{c} \otimes F^{'}$.

In practice, the predicted skeleton map~$F_{sm}$ is downsampled to obtain multi-scale maps~(\ie{1/4, 1/8, 1/16}), that are transferred to the regular stream~(the thick yellow arrow in Fig.~\ref{fig:box}~(a)). And then, the extracted feature map~(\eg{$C1$}) and the corresponding scale skeleton map are used as the input of the Skeleton Attention Module.
Note that the proposed Skeleton Attention Module is shared for the high-level and the low-level feature maps~( \ie $C1, C2, C3$ ).

\begin{figure}[t]
\begin{center}
\includegraphics[width=1\textwidth]{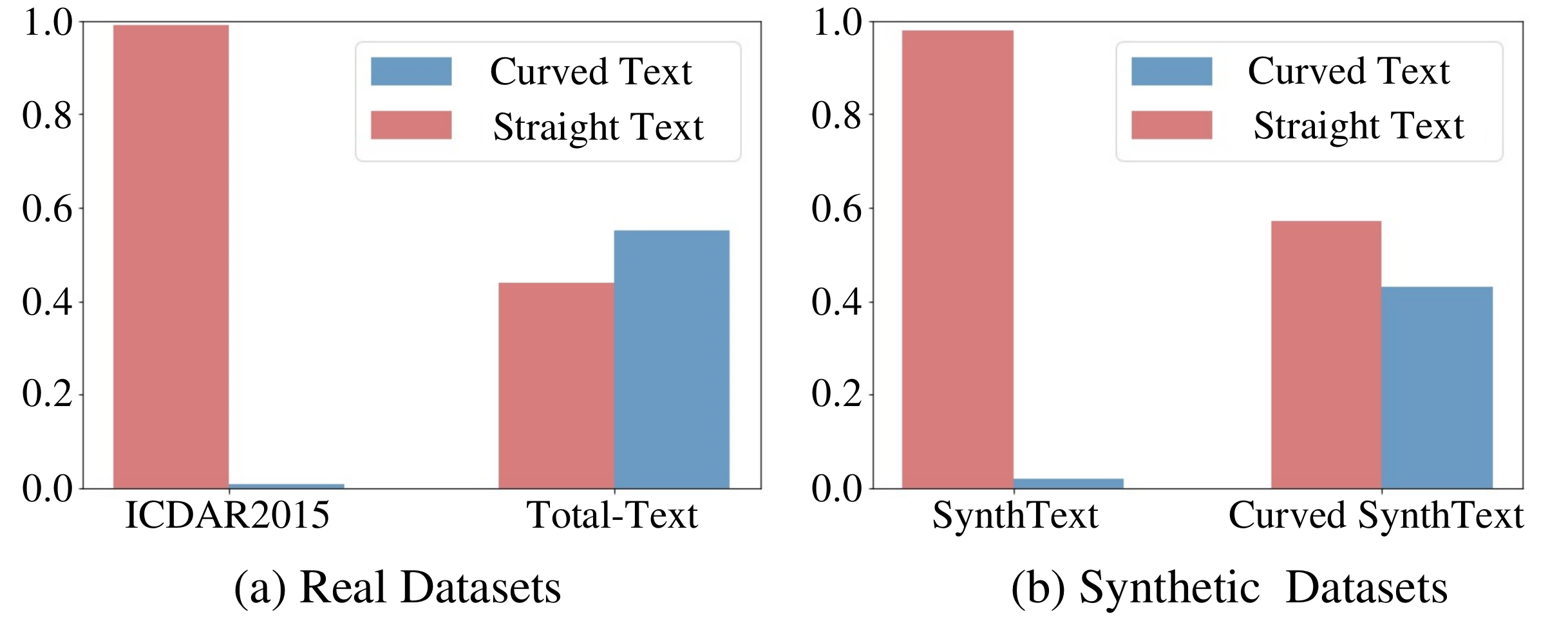}
\caption{The shape distribution difference between synthetic data and real data causes serious domain gap.}
\label{distribution}
\end{center}
\end{figure}

\textbf{Skeleton Attention Map and Loss.} Given an input sample $(x_i,y_i)  \in  \{(x_1,y_1), (x_2,y_2),...(x_n,y_n)\}$, where $x_i$ and $y_i$ denote the $i$-th image and its labels. 
We use $x_i^k$ to denote the $k$-th pixel of the $i$-th training image, with $y_i^k $= $0$ for the background and $y_i^k$ = $1$ for the text pixel.
To learn stronger representation of text skeleton, we define the text skeleton ground-truth as a soft label. For the $k$-th pixel in the text region of $i$-th image, 
we first calculate the shortest distance $d_i^k$ between the $k$-th pixel to its nearest background pixel, 
and then the value $p_i^k$ is defined as the soft skeleton label of $k$-th pixel by normalizing $d_i^k$ to $[0,1]$:
\begin{equation}
\small
p_i^k =  \frac{d_i^k}{d_i^{*}}\,,
\label{equation3}
\end{equation}
where $d_i^{*}$ is the maximum value of $\{d_i^k\}$ in the $i$-th image.

Intuitively, the pixels close to the skeleton of the text instance should have a greater value than the boundary pixels (see text skeleton label in Fig.~\ref{fig:box}~(a)).
Since the soft label is a decimal representing the degree of distance, it is incompatible with the commonly binary cross-entropy loss. 
Besides, the L1 and L2 losses are not sensitive to the distance distribution among $[0, 1]$~\cite{ren2015faster}. 
Therefore, to handle the soft label, we modify the cross-entropy loss into a ``soft'' form. 
For a pixel $x_i^k$, 
$p_i^k$ denotes the value of the $k$-th pixel in the ground truth skeleton map. $\mathcal{F}$ indicates neural networks. The soft cross-entropy loss for text skeleton loss in Fig.~\ref{fig:box}~(a) is defined as follows:

\begin{equation}
\mathcal{L}_{ske} = -\sum_{k}^{}log( 1-\left| p_i^k - \mathcal{F}(x_i^k)\right | )\,,
\end{equation}

The whole loss function $\mathcal{L}$ for SASN can be expressed as a weighted sum of the loss for the regular stream $\mathcal{L}_{ce}$~(\ie{the commonly binary cross-entropy loss}) and the loss for the skeleton stream $\mathcal{L}_{ske}$:
\begin{equation}
\mathcal{L} = \mathcal{L}_{ce} + \lambda \mathcal{L}_{ske},.
\end{equation}
where $\lambda$ is set to 2, which balances the importance between $\mathcal{L}_{ske}$ and $\mathcal{L}_{ce}$.

\subsection{Domain Gap for Synthetic Datasets} 
Existing real-world text datasets can be divided two types: straight~(\eg{ICDAR2015~\cite{karatzas2015icdar}}) and curved~(\eg{Total-Text~\cite{totaltext}}),
depending on the shape of the text instance. However, curved synthetic text data are scarce. SynthText~\cite{gupta2016synthetic}, as the most widely used synthetic dataset, does not contain curved text, which leads to a considerable domain gap between synthetic data and real data~( \eg Total-Text ) in curved text distribution~( $0.2\%~v.s.~58\%$ ), as shown in Fig.~\ref{distribution}.
Therefore, we also adopt Curved SynthText~\cite{long2019rethinking}
to align the data distribution with curved dataset.
In this work,
we use SynthText~\cite{gupta2016synthetic} to train SASN for the straight text line and use Curved SynthText~\cite{long2019rethinking} to match the curved text line.

\section{Experiments}

\subsection{Datasets and Experimental Settings}

\textbf{Pure Synthetic Datasets.} 
\emph{SynthText}~\cite{gupta2016synthetic} consist of 800k synthetic images generated by adding variants of multi-oriented text with random fonts, size, and color. \emph{Curved SynthText}~\footnote{https://github.com/PkuDavidGuan/CurvedSynthText}~\cite{long2019rethinking} generates 80m curved texts with character level annotation by revising the text rendering module of the SynthText engine. 

\textbf{Real Datasets.} \emph{ICDAR2015}~\cite{karatzas2015icdar} includes 1,000 training and 500 testing images with quadrilateral annotation. 
\emph{Total-Text}\cite{totaltext} is an English curved text dataset that contains 1,555 images, including 3 different text orientations: horizontal, multioriented, and curved. 
\emph{MSRA-TD500}~\cite{msra} consists of 500 training and 200 testing images for detecting multi-lingual long texts of arbitrary orientation. 
\emph{ICDAR2017-MLT}~\cite{icdar2017mlt} consists of 18,000 images with texts in 9 languages for multi-lingual text detection.
\emph{SCUT-CTW1500}~\cite{Liu2017Detecting} is a challenging dataset for curved text detection, which consists of 1,000 training images and 500 test images, and text instances are largely in English and Chinese.
\emph{ICDAR2019-ArT}~\cite{chng2019icdar2019} is a large-scale arbitrary-shaped text detection benchmark, and its text regions are labeled by adaptive number of vertices. It consists of 5,603 training images and 4,563 test images.
\emph{ICDAR2019-LSVT}~\cite{sun2019chinese} includes 450,000 images with text that are freely captured in the streets, e.g., store fronts and landmarks. 50,000 of them are fully annotated and are split into 30,000 images for the training set and 20,000 images for the testing set. 

\begin{table}[t]
    \footnotesize
    \centering
\begin{subtable}[t]{3.2in}
    \centering
	\setlength{\tabcolsep}{1.3mm}
	\footnotesize
    \begin{tabular}{l|c|c|c|c}
    \whline
	\multirow{2}*{Method} & \multirow{2}*{Synthetic Data}  &\multicolumn{3}{c}{Evaluation on Total-Text/\%}\cr\cline{3-5}  
	~ & ~ & P & R & F \cr\shline \hline
	\hline
    BL  & SynthText  & 77.2 & 73.0 & 75.0  \\
    BL+SA(C1) & SynthText  & 80.3 & 73.6 & 76.8\textcolor[RGB]{96,177,87}{(+1.8)}  \\
    BL+SA(C1\&C2) & SynthText  & 80.1 & 74.1 & 77.1\textcolor[RGB]{96,177,87}{(+2.1)}  \\
    BL+SA(C1\&C2\&C3) & SynthText  & 80.4 & 74.0 & 77.1\textcolor[RGB]{96,177,87}{(+2.1)}  \\
    \hline
    BL  &  $^*$SynthText & 80.5 & 73.2 & 76.7  \\
    BL+SA(C1) & $^*$SynthText & 81.4 & 75.5 & 78.4\textcolor[RGB]{96,177,87}{(+1.7)}  \\
    BL+SA(C1\&C2) & $^*$SynthText & 81.0 & 76.1 & 78.4\textcolor[RGB]{96,177,87}{(+1.7)}  \\
    BL+SA(C1\&C2\&C3) & $^*$SynthText & \textbf{81.7} & \textbf{75.6} & \textbf{78.5}\textcolor[RGB]{96,177,87}{(+1.8)}  \\
    \whline
    \end{tabular}
	
	\label{table:sa}
	\vspace{-1.5mm}
\end{subtable}
	\caption{\textbf{Multi-Scale features with Skeleton Attention}. `SA', `BL', `P', `R', `F' and `$^*$' refer to `Skeleton Attention', `Baseline', `Precision', `Recall', `F-score' and `Curved SynthText~\cite{long2019rethinking}'.  PSENet~\cite{psenet} is adopted as the detector. The gaps of at least ~(\green{+1.7\%}) improvement compared to the baseline are shown in green.}
	\label{table12}
\end{table}

\begin{table}[t]
    \footnotesize
    \centering
    \begin{subtable}[t]{3.2in}
	\centering
    \setlength{\tabcolsep}{1.3mm}
	\footnotesize
    \begin{tabular}{l|c|c|c}
    \whline
	\multirow{2}*{Method} &\multicolumn{3}{c}{Evaluation on Total-Text/\%}\cr\cline{2-4}  
	~  & Precision & Recall & F-score \cr\shline \hline
    BL    & 80.5 & 73.2 & 76.7  \\
    BL+SA(Cha)    & 79.8 & 74.0 & 76.8\textcolor[RGB]{96,177,87}{(+0.1)}  \\
    BL+SA(Cha\&Spa)    & 80.3 & 74.6 & 77.4\textcolor[RGB]{96,177,87}{(+0.7)}  \\
    BL+SA(Cha\&Spa\&Skeleton Map)    & \textbf{81.7} & \textbf{75.6} & \textbf{78.5}\textcolor[RGB]{96,177,87}{(+1.8)}  \\
    \whline
    \end{tabular}
    
	\label{table:data}
	\vspace{-1.5mm}
\end{subtable}
	\caption{\textbf{Combining methods of Skeleton Attention}. `SA', `BL', `Cha' and `Spa'  refer to `Skeleton Attention', `Baseline', `Channel' and `Spatial'.  PSENet~\cite{psenet} is adopted as the detector. In green are the gaps of up to ~(\green{+1.8\%}) improvement than baseline.}
	\label{table6}
\end{table}

\begin{table*}[h]
    \centering
    
\def\x{{$\footnotesize \times$}}
\footnotesize

\begin{tabular}{l|c|c||c|c|c||c|c|c||c|c|c||c|c|c}
    \whline
	\multirow{2}*{Method} &\multirow{2}*{Annotation} & \multirow{2}*{Pre}  &\multicolumn{3}{c}{ICDAR2015/\%}
	&\multicolumn{3}{c}{MSRA-TD500/\%}
	&\multicolumn{3}{c}{ICDAR2017-MLT/\%}
	&\multicolumn{3}{c}{Total-Text/\%}
	\cr\cline{4-15}  
	~ & ~ & ~ & P & R & F & P & R & F& P & R & F& P & R & F \cr\shline \hline
	\hline
	\textit{Strong Supervision} & ~ & ~ & ~ & ~ & ~& ~ & ~ & ~& ~ & ~ & ~& ~ & ~ & ~  \\
    CTPN\cite{tian2016detecting}  & GT & - & 74.2 & 51.6 & 60.9 & - & - & - & - & - & -& - & - & - \\
    SegLink\cite{shi2017detecting} & GT & \checkmark  & 73.1 & 76.8 & 75.0& 86.0 & 70.0 & 77.0 & - & - & -& 30.3 & 23.8 & 26.7  \\
    EAST\cite{zhou2017east} & GT & -  & 80.5 & 72.8 & 76.4 & 81.7 & 61.6 & 70.2 & - & - & -& - & - & - \\
    PixelLink\cite{PixelLink} & GT & - & 82.9 & 81.7 & 82.3 & 83.0 & 73.2 & 77.8 & - & - & -& - & - & - \\
    TextSnake\cite{textsnake} & GT & \checkmark & 84.9 & 80.4 & 82.6& 83.2 & 73.9 & 78.3 & - & - & -& 82.7 & 74.5 & 78.4  \\
    PSENet\cite{psenet} & GT & - & 81.5 & 79.7 & 80.6  & - & - & - & 73.7 & 68.2 & 70.8& 81.8 & 75.1 & 78.3\\
    PSENet\cite{psenet} & GT & \checkmark & 86.9 & 84.5 & 85.7 & - & - & - & - & - & - & 84.0 & 78.0 & 80.9 \\
    EAST $^{\dagger}$ & GT & - & 76.9 & 77.1 & 77.0 & 71.8 & 69.1 & 70.4 & 68.1 & 63.2 & 65.6& - & - & - \\
    EAST$^{\dagger}$ & GT & \checkmark & \textcolor[rgb]{0,0.55,0.55}{82.0} & \textcolor[rgb]{0,0.55,0.55}{82.4} & \textcolor[rgb]{0,0.55,0.55}{82.2} & \textcolor[rgb]{0,0.55,0.55}{77.9} & \textcolor[rgb]{0,0.55,0.55}{76.5} & \textcolor[rgb]{0,0.55,0.55}{77.2}& \textcolor[rgb]{0,0.55,0.55}{70.3} & \textcolor[rgb]{0,0.55,0.55}{62.8} & \textcolor[rgb]{0,0.55,0.55}{66.4}& - & - & - \\
    PSENet$^{\dagger}$ & GT & - & 81.6 & 79.5 & 80.5 & 80.6 & 77.7 & 79.1& 73.1 & 67.3 & 70.1& 80.4 & 76.5 & 78.4 \\
    PSENet$^{\dagger}$ & GT & \checkmark & \textcolor[rgb]{0,0.55,0.55}{86.4} & \textcolor[rgb]{0,0.55,0.55}{84.0} & \textcolor[rgb]{0,0.55,0.55}{85.2} & \textcolor[rgb]{0,0.55,0.55}{84.1} & \textcolor[rgb]{0,0.55,0.55}{85.0} & \textcolor[rgb]{0,0.55,0.55}{84.5} & \textcolor[rgb]{0,0.55,0.55}{72.5} & \textcolor[rgb]{0,0.55,0.55}{69.1} & \textcolor[rgb]{0,0.55,0.55}{70.8}& \textcolor[rgb]{0,0.55,0.55}{83.4} & \textcolor[rgb]{0,0.55,0.55}{78.1} & \textcolor[rgb]{0,0.55,0.55}{80.7} \\

    \hline
    \hline
    \textit{Weakly Supervision} & ~ & ~ & ~ & ~ & ~& ~ & ~ & ~& ~ & ~ & ~& ~ & ~ & ~\\
    EAST & b-GT & - & 65.8 & 63.8 & 64.8  & 40.5 & 31.1 & 35.2& 66.3 & 59.6 & 62.8& - & - & -\\
    
    EAST+PF & PL & - & 77.8 & 78.2 & 78.0  & 71.3 & 70.2 & 70.7& 67.3 & 64.1 & 65.7 & - & - & -\\
    EAST & b-GT & \checkmark & 70.8 & 72.0 & 71.4& 48.3 & 42.4 & 45.2& 67.2 & 60.1 & 63.5& - & - & -  \\
    EAST+PF & PL & \checkmark & \textbf{81.3} & \textbf{82.2} & \textbf{81.8} & \textbf{77.4} & \textbf{75.5} & \textbf{76.4}& \textbf{67.6} & \textbf{64.9} & \textbf{66.3}& - & - & - \\ 
    \hline
    PSENet & b-GT & - & 70.2 & 69.1 & 69.6  & 47.2 & 36.9 & 41.4& 67.2 & 61.4 & 64.2 & 46.5 & 43.6 & 45.0\\
    
    PSENet+PF & PL & - & 82.9 & 77.6 & 80.2 & 80.3 & 77.5 & 78.9& 72.4 & 69.3 & 70.8 & 81.7 & 75.6 & 78.5 \\
    PSENet & b-GT & \checkmark & 72.7 & 74.3 & 73.5 & 47.5 & 39.5 & 43.1& 66.4 & 63.1 & 64.7& 51.9 & 47.5 & 49.6 \\
    PSENet+PF & PL & \checkmark & \textbf{86.8} & \textbf{84.2} & \textbf{85.5} & \textbf{84.4} & \textbf{84.7} & \textbf{84.5}& \textbf{73.8} & \textbf{67.7} & \textbf{70.6}& \textbf{82.6} & \textbf{78.4} & \textbf{80.5}
    \\ \whline
\end{tabular}
	\caption{\textbf{The results of Polygon-free on ICDAR2015, MSRA-TD500, ICDAR2017-MLT, Total-Text}. $\dagger$ refers to our testing performance. The `GT', `b-GT' and `PL' refer to the 'Ground Truth', `Upright Bounding Box Annotation of Ground Truth' and `Pseudo Label from SASN', respectively. `P', `R', `F' and “Pre” refer to `Precision', `Recall', `F-score' and `pretraining on SynthText', respectively. In \textcolor[rgb]{0,0.55,0.55}{green}~(strong supervision) and in \textbf{bold}~(Polygon-free) are highlighted for comparison.}
	\label{GSG}
\end{table*}

\textbf{Implementation Details.} 
All of the experiments use the same strategy: (1) Training SASN with Curved SynthText based on character annotation, and generating the pixel-level pseudo label on real data based on upright bounding box annotation. (2) Training the detectors~(\ie{ EAST, PSENet, DB}) with the pseudo label. It is worth mentioning that the SASN only needs to train once on Curved SynthText, and we use the same weight to generate pseudo labels on all real benchmarks. Additionally, for all datasets, the threshold for the pseudo-labels generation was the traditional value~(\ie{0.5}). The stochastic gradient descent(SGD) optimizer is adopted with a momentum of 0.9 and a weight decay of 0.0005. The batch size is set to 8 per GPU. The learning rate is initialized to 0.02 and decayed with the power of 0.9 for 16 epochs. During training and inference, the crop images are resized to a resolution of 128$\times$128.  PSENet~\cite{psenet}, EAST~\cite{zhou2017east} and DB~\cite{liao2020real} are adopted as the base detectors because of their popularity. In the PSENet, EAST and DB experiments, all settings follow the original reports.

\subsection{Ablation Study}
Here, we conduct two groups of ablation experiments to analyze PF. The ablation study for synthetic data is in supplementary materials

\textbf{Multi-Scale features with Skeleton Attention.} Tab.~\ref{table12} gives the ablation study about applying the Skeleton Attention on different scale features. Because SA can be applied at different levels for the regular stream, we vary the number of fusion levels from $C1$ to $C3$ to evaluate the relationship. The method that fuses SA with all level features shows the most significant improvement from $76.7\%$ to $78.5\%$. Regardless of the used synthetic data and experimental setting, Skeleton Attention can obtain up to $2\%$ improvement compares to the baseline without the attention.

\textbf{Combining methods of Skeleton Attention.}
Tab.~\ref{table6} shows the impact of three submodules: channel attention, spatial attention, and skeleton attention. It is clear that the improvement due to channel attention is limited~(\ie{$+0.1\%$}), but its spatial attention counterpart contributes to better gain~(\ie{$+0.7\%$}), the main reason for this may be that spatial information is more important than semantic information for text segmentation task. Another important contribution is the soft attention weight map~(\ie{Skeleton Map}), the performance achieves further improvement~(\ie{$+1.8\%$}) after using the Skeleton Map. This is in line with our expectations, and we argue that the text skeleton is vital because of its high representation power.



\begin{table}[h]
    \centering
    \def\x{{$\footnotesize \times$}}
\footnotesize
\begin{tabular}{l|c|c|c|c}
    \whline
	\multirow{2}*{Method} &\multirow{2}*{Annotation}   &\multicolumn{3}{c}{ICDAR2019-LSVT/\%}\cr\cline{3-5}  
	~ & ~ & P & R & F \cr\shline \hline
	\hline
	\textit{Strong Supervision} & ~ & ~ & ~ & ~   \\
    EAST$^{\ast}$ & GT  & 71.7 & 77.6 & 74.5\\
    PSENet$^{\dagger}$ & GT &  80.9 & 74.2 & 77.4\\
    \hline\hline
    \textit{Weakly Supervision} & ~ & ~ & ~ & ~\\
    EAST & b-GT &60.3 & 52.2 & 56.0  \\
    EAST+PF & PL &73.0 & 76.1 & 74.5\textcolor[RGB]{96,177,87}{(+18.5)}  \\
    PSENet & b-GT & 62.1 & 53.2 & 57.3  \\
    PSENet+PF & PL & \textbf{80.1} & \textbf{75.3} &  \textbf{77.6}\textcolor[RGB]{96,177,87}{(+20.3)}  \\
     \whline
\end{tabular}

	\caption{\textbf{The results on ICDAR2019-LSVT}. $\ast$ are collected from \cite{sun2019chinese}. `GT', `b-GT' and `PL' refer to 'Ground Truth', `Upright Bounding Box Annotation' and `Pseudo Label from SASN', respectively. `P', `R' and `F' refer to `Precision', `Recall' and `F-score'. The gaps of at least ~(\green{+18.5\%}) improvement after using BF are shown in green.}
	\label{ICDAR2019-LSVT}
	\vspace{-2mm} 
\end{table}

\subsection{Experiments on Scene Text Detection}
In this section, we present the experiments for Polygon-free on seven datasets, and ICDAR2019-Art and CTW1500 are given in the supplementary material.

\vspace{-3mm}
\subsubsection{Quadrilateral-type datasets}
Tab.~\ref{GSG} lists the experimental results for various methods on the ICDAR2015, ICDAR2017 and MSRA-TD500 datasets, and the ICDAR2019-LSVT is shown in Tab.~\ref{ICDAR2019-LSVT}.

For \emph{ICDAR2015}~\cite{karatzas2015icdar}, PSENet~\cite{psenet} using pseudo label achieves \textbf{almost the same performance}~($85.5\%$ v.s. $85.2\%$) with that using ground truth, proving the high quality of the pseudo label generated by PF. By contrast, direct training the detector with upright bounding box annotation obtains an unsatisfactory F-score~($73.5\%$), with a performance gap of more than $10\%$. EAST~\cite{zhou2017east} shows similar performance~($81.8\% vs. 82.2\%$) to that of PSENet. Moreover, without pretraining on SynthText, the F-score~($78.0\%$) obtained using the pseudo label shows a $1.0\%$ improvement over that~($77.0\%$) of using ground truth. 

For \emph{MSRA-TD500}~\cite{msra}, annotations are provided at the line level, including the spaces between the words in the box. Therefore, bounding box annotation on MSRA-TD500 usually contains a large background, causing poor performance~($35.2\%$ for EAST and $41.4\%$ for PSENet). In this case, the pseudo label from SASN still obtains almost the same performance~($70.7\%$ for EAST and $78.9\%$ for PSENet without pre-trained model) as that~($70.4\%$ for EAST and $79.1\%$ for PSENet) of polygon annotation. With the pre-trained model, a similar result is obtained, and PSENet with pseudo label~($84.5\%$) achieves a huge improvement~($\textbf{+41.4}\%$) compared to training directly with upright bounding box~($43.1\%$).

For \emph{ICDAR2017-MLT}~\cite{icdar2017mlt}, the performance is similar to ICDAR15, and pseudo label shows strongly improved performance compared to that using ground truth. The difference is that the upright bounding box also leads to relatively good performance~($62.8\%$ for EAST and $64.2\%$ for PSENet). The main reason for this is that the texts in ICDAR2017-MLT have a small tilting angle and size.

Tab.~\ref{ICDAR2019-LSVT} gives the results for \emph{ICDAR2019-LSVT}~\cite{sun2019chinese}. With pseudo label from SASN, PSENet achieves $77.6\%$ F-score, $\textbf{20.3\%}$ better than training directly with upright bounding box annotation, and almost the same as that obtained using original polygon annotation~($77.4\%$). A similar case with $\textbf{18.5\%}$ performance improvements for EAST. For large-scale ICDAR2019-LSVT with \textbf{50k} images, the excellent performance has great significance for reducing the annotation cost, and for the first time, the great potential of box-supervised text detection is revealed.

\subsubsection{Curved-type dataset}
\begin{figure}[!t]
\begin{center}
\includegraphics[width=0.97\textwidth]{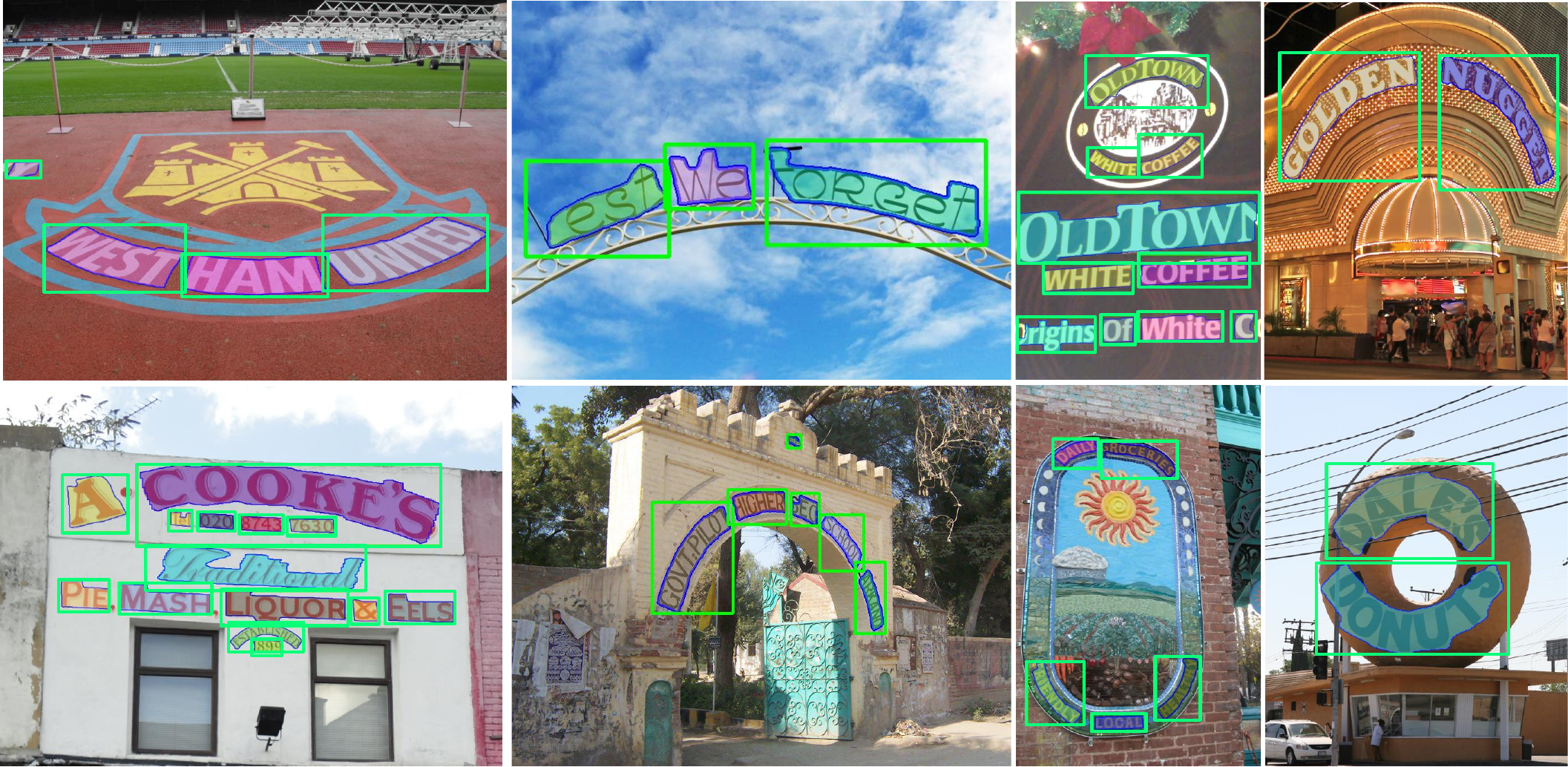}
\caption{\textbf{Visualization for the pseudo labels from Polygon-free}. The images are from Total-Text~\cite{totaltext}.}
\label{example}
\end{center}
\vspace{-5mm}
\end{figure}

For \emph{Total-Text}~\cite{totaltext}, Tab.~\ref{GSG} lists the experimental results. The annotation on Total-Text is complex and polygonal in shape. The \textbf{great performance}~($80.5\%$) of Polygon-free further demonstrates the significance of our work, and Fig.~\ref{example} provides some visualization of ground truth and pseudo label. Similar to MSRA-TD500, the upright bounding box annotation on Total-Text also contains plenty of backgrounds, causing poor performance~($45.0\%$ and $49.6\%$). The use of the pseudo label generated by SASN can still achieve excellent performance~($78.5\%$ and $80.5\%$) with improvements of $\textbf{33.5\%}$ and $\textbf{30.9\%}$. 



\begin{figure}[!t]
\begin{center}
\includegraphics[width=0.97\textwidth]{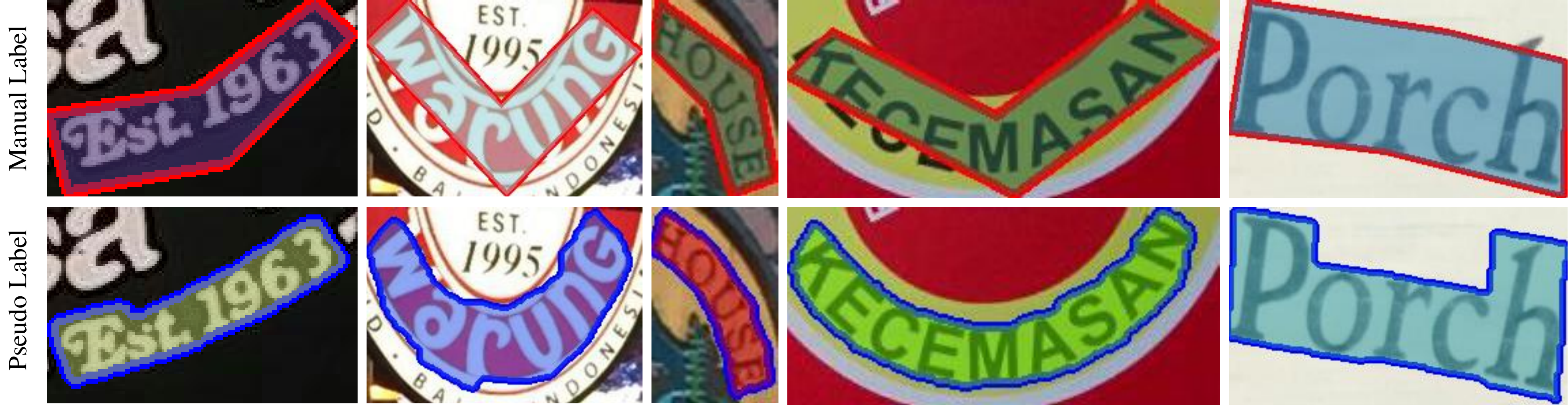}
\caption{\textbf{Comparison of pseudo label and manual label on Total-Text~\cite{totaltext}}. Pseudo label is more smooth than the manual label.}
\label{pseudolabel}
\end{center}
\vspace{-5mm}
\end{figure}

\begin{table}[t]
    \centering
    \begin{subtable}[t]{3.2in}
	\centering
    \setlength{\tabcolsep}{1.3mm}
	\footnotesize
    \begin{tabular}{l|c|c|c|c|c}
    \whline
	Datasets & Cited & Methods& F1$^{\blacktriangle }$/\% & F1$^{\bigstar }$/\% & F1$^{\dagger}$/\% \cr\shline \hline \hline
    \multirow{2}*{ICDAR2015} & \multirow{2}*{624} & EAST & 64.8 & 78.0~(\green{+13.2}) & 76.4  \\ 
    ~ & ~ & PSENet &  73.5  & 85.5~(\green{+12.0})  & 85.7  \\
    MSRA-TD500 & 630 & EAST & 35.2  & 70.7~(\green{+35.5})  & 70.2  \\
    ICDAR2017MLT & 128 & PSENet & 64.2  & 70.8~(\green{+6.6})  & 70.8  \\
    \multirow{2}*{Total-Text} & \multirow{2}*{142} & PSENet & 49.6  & 80.5~(\green{+30.9})  & 80.9  \\
    ~ & ~ & \textbf{DB} &  49.1  & 84.5(\green{+35.4})  & 84.7  \\
    CTW1500 & 105 & PSENet & 47.6  & 81.5(\green{+33.9})  & 82.2  \\
    \textbf{ICDAR2019LSVT} & 6 & PSENet & 57.3  & 77.6(\green{+20.3})  & 77.4  \\
    \textbf{ICDAR2019ArT} & 13 & PSENet & 46.6  & 69.2(\green{+22.6})  & 69.5  \\
    \whline
    \end{tabular}
    
	\label{table:data}
\end{subtable}
	\caption{\textbf{Weakly supervision \emph{v.s}. Strong supervision}. $\blacktriangle$, $\bigstar$ and $\dagger$ refer to `Upright Bounding Box Annotation', `Polygon-free(ours)' and `Original Paper Report or our testing result~(strong supervision)'. In blue are the gaps of at least ~(\green{+6.6\%}) improvement after using PF.}
	\vspace{-2mm} 
	\label{supervision}
\end{table}

\subsection{Discussion}
\textbf{Weakly supervision \emph{v.s}. Strong supervision.} To further present the effectiveness of the Polygon-free, we summarized the results concerning three detectors~(\ie{PSENet~\cite{psenet}, EAST~\cite{zhou2017east} and DB~\cite{liao2020real}}) and seven datasets~(\ie{ICDARs, Total-Text, CTW1500 and MSRA-TD500}) to Table~\ref{supervision}. As a \textbf{plug-and-play} weakly-supervised approach, the F1 of PF can achieve $6.6\%$ $\sim$ $35.5\%$ improvements than directly training with upright bounding box~(two points), almost equal to strong-supervised methods on all datasets. This means that the proposed method can be directly applied for industry application with a little loss~(\ie{$< 0.5\%$}) of performance. The competitive performance of PF proves the practicability efficiency of the pseudo label. Fig.~\ref{example} gives some visualization of the pseudo label.

\textbf{Pseudo Label \emph{v.s}. Manual Label.}
With the ideal condition, compared with the pseudo label, the manual label is more accurate. However, since many factors such as disagreement from multiple annotators, there are many label errors or low-quality manual labels. As shown in Fig.~\ref{pseudolabel}~(the first row), manual labels even can not contain the whole text region, which will cause serious inference for the following recognition task. As the number of data increases, it is more difficult to assure data quality, and even high-quality datasets are likely to contain incorrect labels. By contrast, in these cases, the pseudo label is more \textbf{smooth} for suiting each character, which can obtain a better performance via supervising the network. Besides, we argue that Polygon-free can be used in other tasks to boost their efficiency. For example, designing an automatic annotation tool with PF, more smooth and high-quality annotation can be obtained by handling some bad cases at a low cost.   





\section{Conclusion}
In this paper, we present a simple but surprisingly effective and practical system termed Polygon-free, in which most existing polygon-based text detectors are trained with only upright bounding box annotations. The core contribution is to transfer knowledge from synthetic data to real data to enhance the supervision information via a skeleton attention segmentation network. The experiments showed that our method achieves almost the same performance as that of strong supervision while saving huge annotation cost~(\eg{$80\%$ + data cost on Total-Text}), which can provide a new perspective for weakly supervised text detection and save much money for for real-world application.

{\small
\bibliographystyle{ieee_fullname}
\bibliography{egbib}
}

\end{document}